\documentclass[10pt, conference, letterpaper]{IEEEtran}

\usepackage[
backgroundcolor = White,textwidth=\marginparwidth]{todonotes}

\usepackage[markup=underlined]{changes}

\usepackage[utf8]{inputenc} 
\usepackage[T1]{fontenc}    
\usepackage{url}            
\usepackage{booktabs}       
\usepackage{amsfonts}       
\usepackage{nicefrac}       
\usepackage{microtype}      

\usepackage{graphicx}
\usepackage{color}
\usepackage{rotating}
\usepackage{tabularx}
\usepackage{pdflscape}
\usepackage{amsmath,amsthm,amssymb}
\usepackage{algorithm,algorithmic}
\usepackage{bbm,dsfont}
\usepackage{caption,subcaption}
\usepackage{wrapfig}
\usepackage{cancel}
\usepackage{enumerate, cases}
\usepackage{thmtools,thm-restate}
\usepackage{mathtools}

\usepackage[none]{hyphenat}
\usepackage{multirow}
\usepackage{makecell}
\usepackage{setspace}
\usepackage{pifont}

\allowdisplaybreaks

\usepackage{dblfloatfix}
\usepackage{array}
\usepackage{enumitem}

\allowdisplaybreaks

\newcommand{\Alg}{\mathcal{A}}

\newcolumntype{C}[1]{>{\centering}m{#1}}
\newcommand{\EE}[1]{\mathbb{E}\left[#1\right]}

\newcommand{\Prob}[1]{\mathbb{P}\left\{#1\right\}}
\newcommand{\Regret}{\mathcal{R}}

\newcommand{\one}[1]{\mathds{1}_{\left\{#1\right\}}}

\newcommand{\btheta}{\boldsymbol{\theta}}

\newcommand{\WD}{\mathrm{WD}}

\newcommand{\TWD}{\Theta_{\WD}}

\newcommand{\MD}{\Theta_{\mathrm{MD}}}

\newcommand{\Yti}{Y_t^i}
\newcommand{\Ytj}{Y_t^j}

\newcommand{\Yt}{Y_t}
\newcommand{\Yi}{Y^i}
\newcommand{\Yj}{Y^j}
\newcommand{\Yis}{Y^{i^\star}}

\newcommand{\opij}{\hat{p}_{ij}^o}
\newcommand{\opji}{\hat{p}_{ji}^o}
\newcommand{\opijts}{\hat{p}_{ij}^{ts}}
\newcommand{\opijkl}{\hat{p}_{ij}^{kl}}
\newcommand{\opijucb}{\hat{p}_{ij}^{ucb1}}

\newcommand{\Dij}{\mathcal{D}_{ij}(t)}
\newcommand{\Nij}{\mathcal{N}_{ij}(t)}
\newcommand{\Dijm}{\mathcal{D}_{ij}(t-1)}
\newcommand{\Nijm}{\mathcal{N}_{ij}(t-1)}

\newcommand{\bd}{\boldsymbol{D}}
\newcommand{\bcost}{\boldsymbol{c}}

\theoremstyle{plain}

\newtheorem{defi}{Definition}

\begin{document}

	\title{
		Unsupervised Online Feature Selection for Cost-Sensitive Medical Diagnosis
		}
	
	\author{
		Arun Verma, Manjesh K. Hanawal, and Nandyala Hemachandra \\
		Industrial Engineering and Operations Research\\
		Indian Institute of Technology Bombay,
		Mumbai, India - 460076 \\
		\{v.arun, mhanawal, nh\}@iitb.ac.in
	}
	
	\maketitle
	
	\begin{abstract}\
		In medical diagnosis, physicians predict the state of a patient by checking measurements (features) obtained from a sequence of tests, e.g., blood test, urine test, followed by invasive tests.  As tests are often costly, one would like to obtain only those features (tests) that can establish the presence or absence of the state conclusively. Another aspect of medical diagnosis is that we are often faced with unsupervised prediction tasks as the true state of the patients may not be known. Motivated by such medical diagnosis problems, we consider a {\it Cost-Sensitive Medical Diagnosis} (CSMD) problem, where the true state of patients is unknown. We formulate the CSMD problem as a feature selection problem where each test gives a feature that can be used in a prediction model. Our objective is to learn strategies for selecting the features that give the best trade-off between accuracy and costs. We exploit the `Weak Dominance' property of problem to develop online algorithms that identify a set of features which provides an `optimal' trade-off between cost and accuracy of prediction without requiring to know the true state of the medical condition. Our empirical results validate the performance of our algorithms on problem instances generated from real-world datasets.
	\end{abstract}
	

	\IEEEpeerreviewmaketitle
	\bstctlcite{IEEEexample:BSTcontrol} 	

	\section{Introduction}
	\label{sec:introduction}

In medical diagnosis,  physicians identify the presence of a particular medical state by performing a set of tests that give measurements related to symptoms of the state. For example, to identify diabetes, tests like A1C, FPG, or RPG  are performed to measure blood sugar levels. Test outputs can be treated as features that physicians can use to predict the presence or absence of a medical state. More features may aid physicians in making better predictions. However, obtaining them increases the cost of diagnosis: this can be actual monetary cost, or in terms of the time taken to get the results, or the side effects it has on the human body. In such cases, one has to decide what is the set of features to acquire that gives the best trade-off between the costs and the prediction accuracy. In addition to this, the medical diagnosis has a peculiar structure -- the tests need to be performed in a sequential order. For example, invasive tests are performed after a urine/blood test. In technical terms, it requires that the tests form a cascade, and the total cost increases with the number of tests. The other challenging issue in the medical diagnosis is that the true state of patients may not be known, and hence, the effectiveness of tests is unknown, making the learning essentially unsupervised. We term such problems as \textit{\underline{C}ost-\underline{S}ensitive \underline{M}edical \underline{D}iagnosis} (CSMD).

\begin{figure}[H]
	\centering
	\includegraphics[scale=.37]{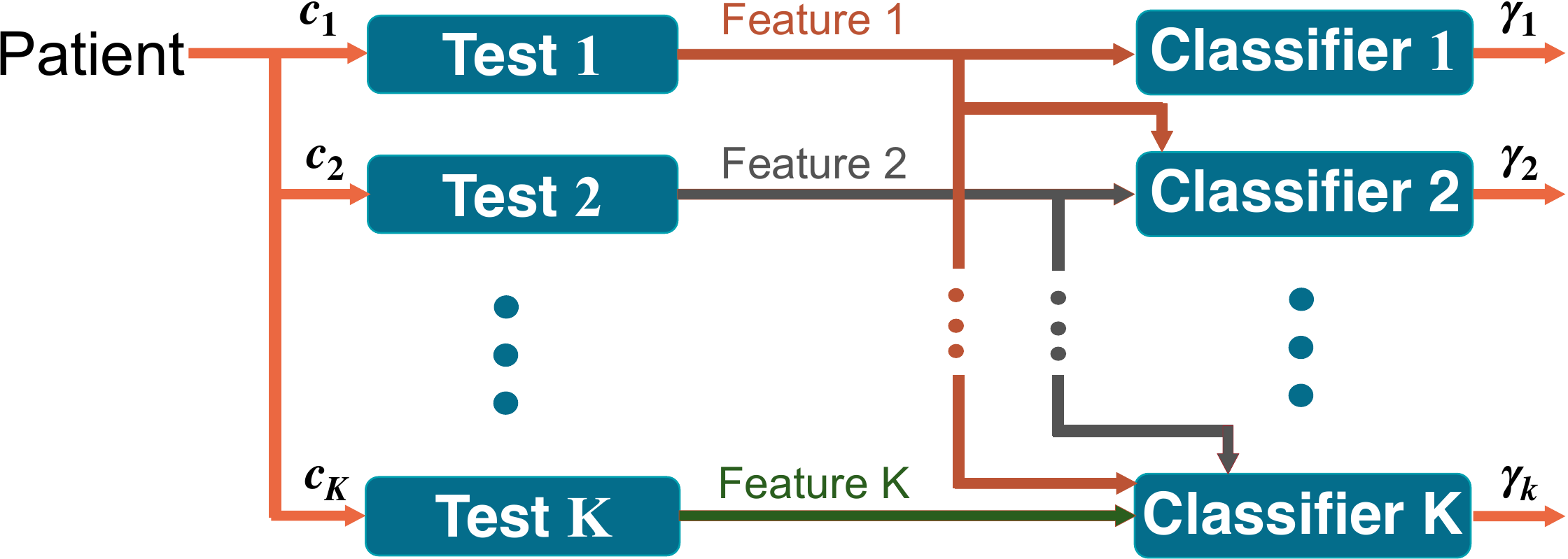}
	\caption{A cascade of tests where Classifier $k$ uses features obtained from tests $1$ to $k$ to make predictions. Classifier $k$ has error-rate $\gamma_k$ and test $k$ has cost $c_k$.}
	\label{fig:CSMD}
\end{figure}

In the CSMD setup, it is assumed that the tests form a cascade where tests are performed in a fix sequence. Each test outputs a feature, and collection of features are used to predict the presence of a medical state by applying a classifier on them.  The error-rate of the classifiers are unknown, and the cost associated with a classifier is the sum of the costs of the tests used to obtain its input features.  The loss of a classifier is defined as the sum of its error-rate and cost of input features. The aim is to find a classifier that has the smallest mean loss, i.e., a set of tests/features that gives the best trade-off between cost and error-rate. The CSMD setup is shown in Figure \ref{fig:CSMD}.

The CSMD setup is unsupervised as the medical state of the patients is assumed to be unknown, and hence the error-rates of classifiers cannot be computed. Clearly, learning is not feasible unless we exploit the problem structure, which reveals some information about the state either explicitly or implicitly. In \cite{AISTATS17_hanawal2017unsupervised} and \cite{AISTATS19_verma2019online}, it is shown that it is possible to get a good estimate of the error-rates in such unsupervised setting provided the problem has some structure. Specifically, it is shown that if the problem satisfies strong dominance (SD) or weak dominance (WD) property, learning of the optimal classifier is possible. The SD property requires that if a classifier's prediction is correct, then all the classifiers that follow it in the cascade also have correct predictions. The WD property relaxes this constraint on predictions and allows errors in some instances from better tests. We exploit these properties to develop algorithms.

The CSMD problem is similar to a feature selection problem \cite{guyon2003introduction} in machine learning, where each test output corresponds to knowing a feature (condition) and then inferring the state of the patient from the obtained features. Since patients arrive sequentially, the identification of the optimal feature set becomes an online feature selection problem. We adopt solution techniques such as Thompson Sampling (TS) \cite{COLT12_agrawal2012analysis} and Upper Confidence Bound (UCB) indices: UCB1 \cite{JMLR02_auer2002using} and kl-UCB \cite{COLT11_garivier2011kl}, available in the multi-armed bandit literature \cite{BOOK_lattimorebandit} to address the problem of online feature selection in CSMD.

Our contributions are summarized as follows:
\begin{itemize}
    \item In Section \ref{sec:cascade}, we assume that the tests form a cascade and are performed in a fix sequence. We propose TS and kl-UCB based algorithms named \ref{alg:CSMD_TS} and \ref{alg:CSMD_KL} respectively for such online feature selection in CSMD.
	
	\item We relax cascade assumption in Section \ref{sec:subset}, where any subset of tests can be selected. For this setup, we develop the algorithms named \ref{alg:CSMD_CTS} (TS based) and \ref{alg:CSMD_ESCB} (UCB based).
	
	\item In Section \ref{sec:experiments}, we demonstrate the performance of our algorithms on instances derived from real-world datasets. Our experimental results show that TS based algorithms always perform better than UCB based algorithms.
\end{itemize}

\subsection{Related Work}
The multi-armed bandits is first introduced for designing optimal treatment allocation in medical science by \cite{BIOMETRIKA33_thompson1933likelihood}. Since then, many applications to medical sciences such as dynamic treatment regime \cite{EJS14_laber2014dynamic}, treatment strategies for epilepsy \cite{IJNS09_pineau2009treating} or cancer \cite{AAAI15_durand2015adaptive}, have been proposed using the decades of research in the field of multi-armed bandits. 

Online Feature Selection problem is first explored in \cite{wang2014online}, where the features at every iteration are selected such that the number of mistakes in classification is minimized. \cite{durand2014thompson} use Thompson Sampling based algorithm for online feature selection. However, the literature on learning optimal actions in the unsupervised online setting is limited. \cite{AISTATS17_hanawal2017unsupervised} and \cite{AISTATS19_verma2019online} propose an unsupervised online learning setting and develope conditions when the optimal action/test can be identified. We build on this work and apply it to solve online feature selection in CSMD. While \cite{AISTATS17_hanawal2017unsupervised} and \cite{AISTATS19_verma2019online} focus on UCB based algorithms, we also propose TS based algorithms. Further, \cite{AISTATS17_hanawal2017unsupervised} and \cite{AISTATS19_verma2019online} require that features are selected sequentially (cheaper feature followed by expensive features) from a cascade, whereas we relax this assumption by allowing to select any subset of features.

	\section{Problem Setting}		
	\label{sec:problemSetting}

We pose a CSMD problem as a multi-armed bandits problem where loss from arms is not directly observed. Specifically, each test's output is treated as a feature, and we have different classifiers for a different set of features. As the feature selection is sequential (relaxed later), the $k$-th classifier corresponds to the features $\{1,2,\ldots,k\}$. The output of the classifier is $1$ if the result is positive; otherwise, it is $0$. The classifier represents an arm in the corresponding multi-armed bandits problem. A problem instance $\btheta$ is specified by a pair $ (\bd,\bcost)$, where $\bd$ is a distribution over the $K+1$ dimensional hypercube corresponding to the true state and the outputs of the $K$ arms, and $\bcost$ is a $K$-dimensional non-negative valued vector of costs, where $K$ is the number of features. $\bcost$ is known to the learner from the start, while $\bd$ is unknown. Following the terminology of multi-armed bandits, henceforth, we refer to the classifiers as arms and their predictions as feedback. The learner-environment interaction is given in Algorithm \ref{alg:CSMD}.
\begin{algorithm}[!h]
	\caption{CSMD Learning Problem for instance $(\bd, \bcost)$}
	\label{alg:CSMD}	For each round $t$: 
	\begin{enumerate}
		\item \textbf{Environment} generates a $K+1$-dimensional binary vector $(Y_t,Y_t^1,\dots,Y_t^K) \in D$ where $\EE{\Yti \ne \Yt} = \gamma_{i}$, $Y_t^i$ is the output of arm $i$, while $Y_t$ is the unknown true state to be guessed by the learner. 	
		\item \textbf{Learner} chooses an arm $I_t \in \{1,2, \ldots, K\}$		 
		\item \textbf{Feedback and Loss:} The learner observes $Y_t^1,\dots,Y_t^{I_t}$ and incurs loss $\lambda_{I_t}C_{I_t} +\one{Y_t \ne Y_t^{I_t}}$		
	\end{enumerate}
\end{algorithm}

The cost of acquiring feature $k \in [K]$ is denoted by $c_k \geq 0$  where $[K] \doteq \{1,2, \ldots, K\}$. Due to sequential selection, the total cost of choosing arm $I_t$ in round $t$ is $C_{I_t} \doteq c_1 + \dots + c_{I_t}$.  After selecting arm $I_t$, a learner observes $\{Y_t^i\}_{i=1}^{I_t}$ where $Y_t^i$ represents the feedback of arm $i$. The {\it expected total cost} incurred in round $t$ is $\EE{\lambda_{I_t} C_{I_t} + \one{Y_t^{I_t} \ne \Yt}}=\lambda_{I_t}C_{I_t} +\gamma_{I_t}$ where $\gamma_{I_t} \doteq \Prob{\Yt^{I_t} \ne \Yt}$ is the error-rate of arm $I_t$ and $\lambda_{I_t}$ normalizes error-rate and cost of using arm $I_t$. Without loss of generality (WLOG), we set $\lambda_{i}= 1$ for all $i \in [K]$. The optimal arm minimizes expected total cost which is given by 
\begin{equation}
	\label{eq:OptimalTest}
	i^\star \doteq \max\left\{s: s= \arg\min\limits_{j \in [K]} \left(C_{j} +\gamma_j \right) \right\}.
\end{equation}

The goal of the learner is to find optimal arm in every round such that the cumulative loss is minimized. Specifically, we measure the performance of a policy that selects arm $I_t$ in round $t$ and over $T$ rounds in terms of expected cumulative (pseudo-)regret given by
\begin{align}
	\label{eq:regret}
	\mathbb{E}[\Regret_T] = \sum_{t=1}^T \left[ \left( C_{I_t} +\gamma_{I_t} \right) - \left( C_{i^\star} +\gamma_{i^\star} \right) \right].
\end{align}
A good policy should have sub-linear expected regret, i.e., $\mathbb{E}[\Regret_T]/T \to 0$ as $T\to \infty$, where the expectation is over the environment's randomness and in the choice of $I_t$ induced by the environment. If the regret is sub-linear, the learner collects almost as much reward in expectation in the long run as an oracle that knew the optimal arm from the beginning. 

Let $\MD$ be the set of all CSMD problems. Given a subset $\Theta\subset \MD$, we say that $\Theta$ is \emph{learnable}  if there exists a learning algorithm $\Alg$ such that for any $\theta\in \Theta$, the expected regret $\EE{ \Regret_T(\Alg,\theta) }$ of algorithm $\Alg$ on instance $\theta$ is sub-linear. A subset $\Theta$ is said to be a \emph{maximal learnable} problem class if it is learnable and for any subset $\Theta^\prime \subset \MD$ that contains $\Theta$ is not learnable. 

Now we define Weak Dominance (WD) property of problem instance that makes the learning of the optimal arm possible.
\begin{defi}
	\label{def:WD} 
	Let $i^\star$ be an optimal arm. Then an instance $\theta \in \MD$  is said to satisfy \emph{weak dominance property} if
	\begin{align}
		\label{eq:WD}
		\xi(\theta) := \min_{j>i^\star} \Big\{C_j - C_{i^\star} - \Prob{\Yis \ne \Yj} \Big\} > 0
	\end{align}
\end{defi}
\vspace{-.2cm}
Let $\TWD = \left\{\theta \in \MD:\theta \mbox{ satisfies $\WD$ condition}\right\}$ denote the set of instances satisfying the $\WD$ property. \cite{AISTATS17_hanawal2017unsupervised} and \cite{AISTATS19_verma2019online} have shown that the set of problems satisfying the WD property is maximally learnable. Any relaxation of WD property makes the problem unlearnable.

	\section{Learning  with Cascade Structure}
	\label{sec:cascade}

In this section, we focus on the case where features form a cascade and are acquired in a fixed sequence. A feature can be used only after using all previous features in the cascade. Let $i^\star$ denote the optimal arm, i.e., set of optimal features $\{ 1,$ $2, \ldots, i^\star \}$. Then $i^\star$ must satisfy the following inequalities\footnote{The inequalities lead to risk-averse selection criteria, i.e., if two arms have the same total cost, the arm with smaller error-rate will be chosen.}:
\begin{subequations}
	\label{eq:cost_exp_err}
	\begin{align}
	&\forall j<i^\star \,:\, C_{i^\star} - C_j \leq \gamma_j-\gamma_{i^\star} \,, \label{eq:wd1}\\ 
	&\forall j>i^\star \,:\, C_j - C_{i^\star} > \gamma_{i^\star}-\gamma_j \,. \label{eq:wd2}
	\end{align}
\end{subequations}
Due to the unavailability of true state, error rates ($\gamma_j$) cannot be estimated, implying that \eqref{eq:wd1} and \eqref{eq:wd2} do not lead to a sound arm selection criteria. However, if any problem instance satisfies WD property then only optimal arm $i^\star$ lies in the following two sets \cite{AISTATS19_verma2019online}: 
\begin{subequations}
	\label{eq:Sets}
	\begin{align}
		&\mathcal{B}^l= \Big\{i: \forall j<i, C_i - C_j \leq \Prob{\Yi \ne \Yj} \Big\} \cup \{1\}, \label{set:Bl} \\
		&\mathcal{B}^h = \Big\{i: \forall j>i, C_j - C_i > \Prob{\Yi \ne \Yj} \Big\} \cup \{K\}. \label{set:Bh}
	\end{align}
\end{subequations}
The disagreement probabilities $\Prob{\Yi \ne \Yj}$ for all $(i,j)$ pair are unknown (but fixed). Hence, in the round $t$, any algorithm has to estimate the disagreement probability by comparing feedback from the selected arms and replaces $\Prob{\Yi \ne \Yj}$ by its optimistic estimates $\opij(t)$ as follows:
\begin{subequations}
	\label{eq:estimateOptimal}
	\begin{align}
		&\mathcal{\hat{B}}_t^l= \{i: \forall j<i, C_{i}-C_{j} \leq \opji(t) \} \cup \{1\}, \label{set:eBl} \\
		&\mathcal{\hat{B}}_t^h=\{i: \forall j>i;C_{j} -C_{i} > \opij(t) \} \cup \{K\}. \label{set:eBh}
	\end{align}
\end{subequations}
Due to symmetry, $\opij(t) = \opji(t)$ for any $(i,j)$ pair in round $t$. Therefore, it is sufficient for an algorithm to only keep track of $\opij$ for $i<j$. Next, we describe Thompson Sampling \cite{COLT12_agrawal2012analysis} and kl-UCB \cite{COLT11_garivier2011kl} based two online algorithms that check the conditions given in \eqref{eq:estimateOptimal} for selecting the optimal arm.

\subsection{Thompson Sampling based algorithm: \ref{alg:CSMD_TS}}
The Upper Confidence Bound (UCB) is very effective for dealing with the trade-off between exploration and exploitation in bandit problems. UCB has been widely used for solving various sequential decision-making problems. On the other hand, Thompson Sampling (TS) is an online algorithm based on Bayesian updates that chooses an arm to play according to its probability of being the best arm. TS has been shown to be empirically superior in comparison to UCB based algorithms for various MAB problems by \cite{NIPS11_chapelle2011empirical}. TS achieves lower bound for MAB \cite{AAM85_lai1985asymptotically} when rewards of arms have Bernoulli distribution is shown by \cite{ALT12_kaufmann2012thompson}.
\begin{algorithm}[!t]
	\renewcommand{\thealgorithm}{\textnormal{CSMD-TS}}
	\floatname{algorithm}{}
	\caption{Algorithm for \textbf{CSMD} problem using \textbf{TS}} 
	\label{alg:CSMD_TS}
	\begin{algorithmic}[1]
		\STATE Set $ \forall 1 \le i< j \le K: \mathcal{S}_{ij}(1) \leftarrow 1, \mathcal{F}_{ij}(1) \leftarrow 1$
		\FOR{$t=1,2,...$}
			\STATE $\forall 1 \le i< j \le K: \opijts(t) \leftarrow \mbox{Beta}(\mathcal{S}_{ij}(t), \mathcal{F}_{ij}(t))$
			\STATE Compute $\mathcal{\hat{B}}_t^l$ and $\mathcal{\hat{B}}_t^h$ as given in \eqref{set:eBl} and \eqref{set:eBh} by setting $\opij(t) = \opijts(t)$. Set $\mathcal{\hat{B}}_t := \mathcal{\hat{B}}_t^l \cap \mathcal{\hat{B}}_t^h$
			\STATE $I_t\leftarrow \min \left\{ \mathcal{\hat{B}}_t\cup \{K\}\right\}$
			\STATE Select arm $I_t$ and observe $Y_t^1,\dots,Y_t^{I_t}$
			\STATE $\forall 1\le i< j \le I_t:$ update $\mathcal{S}_{ij}(t+1) \leftarrow\mathcal{S}_{ij}(t) + \one{\Yti \ne \Ytj}, \mathcal{F}_{ij}(t+1) \leftarrow\mathcal{F}_{ij}(t) + \one{\Yti = \Ytj}$
		\ENDFOR
	\end{algorithmic}
\end{algorithm}
We develop a Thompson Sampling based algorithm, named \ref{alg:CSMD_TS}, that uses selection criteria given in \eqref{eq:estimateOptimal} to select optimal arm. \ref{alg:CSMD_TS} works as follows. It sets the prior distribution of each pair of arms as the Beta distribution Beta$(1, 1)$, which is same as Uniform distribution on $[0,1]$. $S_{ij}$ represents the number of rounds when a disagreement is observed between arm $i$ and $j$ whereas $F_{ij}$ represents the number of rounds when an agreement is observed. Let $S_{ij}(t)$ and $F_{ij}(t)$ denotes the values of $S_{ij}$ and $F_{ij}$ at the beginning of round $t$. In round $t$, a sample $\opijts$ is independently drawn from Beta$(S_{ij}(t), F_{ij}(t))$ for each pair of arms $(i,j)$ where $1 \le i< j\le K$ (Line $3$). The sets $\hat{\mathcal{B}}_t^l$, $\hat{\mathcal{B}}_t^h$, and their intersection are computed by setting $\opij$ as $\opijts$ (Line $4$). Then \ref{alg:CSMD_TS} selects an arm $I_t$ that satisfies \eqref{set:eBl} and \eqref{set:eBh} (Line $5$). Note that initially $\hat{ \mathcal{B}}_t$ can be empty due to the bad estimates of disagreement probabilities. In such a case, the arm $K$ is selected. After selection of arm $I_t$, the feedback $\{Y^j_t\}_{j=1}^{I_t}$ (Line $6$) is observed which is used to update the values of $S_{ij}(t+1)$  and $F_{ij}(t+1)$ (Line $7$).

\subsection{kl-UCB based algorithm: \ref{alg:CSMD_KL}}
A UCB based algorithm USS-UCB is proposed by \cite{AISTATS19_verma2019online}, which can be adapted for solving the CSMD problem. The kl-UCB algorithm \cite{COLT11_garivier2011kl} satisfies a uniformly better regret bound than UCB1 \cite{JMLR02_auer2002using} and its variants. For Bernoulli rewards, kl-UCB reaches the lower bound of \cite{AAM85_lai1985asymptotically}. Adapting kl-UCB for CSMD, we develop an algorithm named \ref{alg:CSMD_KL}. \ref{alg:CSMD_KL} works as follows. It needs input $a$ which is recommends to set $a = 0$ for optimal performance in practice by \cite{COLT11_garivier2011kl}. 
\begin{algorithm}[H]
	\renewcommand{\thealgorithm}{\textnormal{CSMD-kl}}
	\floatname{algorithm}{}
	\caption{Algorithm for \textbf{CSMD} problem using \textbf{kl}-UCB }
	\label{alg:CSMD_KL}
	\begin{algorithmic}[1]
		\STATE \textbf{Input:} $a \ge 0$
		\STATE Select arm $K$ and observe $Y^1,\dots,Y^K$. Set $\forall 1 \le i< j \le K \hspace{-0.25mm}:\hspace{-0.25mm} \mathcal{D}_{ij}(1) \hspace{-0.5mm}\leftarrow\hspace{-0.5mm} \one{Y^i_1\ne Y^j_1}, \mathcal{N}_{ij}(1) \hspace{-0.5mm}\leftarrow\hspace{-0.5mm} 1$
		\FOR{$t=2,3,...$}
			\STATE $\forall 1 \le i< j \le K:$ compute $\opijkl(t)$ using \eqref{eq:KL_Estimate}	
			\STATE Compute $\mathcal{\hat{B}}_t^l$ and $\mathcal{\hat{B}}_t^h$ as given in \eqref{set:eBl} and \eqref{set:eBh} by setting $\opij(t) = \opijkl(t)$. Set $\mathcal{\hat{B}}_t := \mathcal{\hat{B}}_t^l \cap \mathcal{\hat{B}}_t^h$
			\STATE $I_t\leftarrow \min \left\{ \mathcal{\hat{B}}_t\cup \{K\}\right\}$
			\STATE Select arm $I_t$ and observe $Y_t^1,\dots,Y_t^{I_t}$
			\STATE $\forall 1 \le i< j \le I_t:$ update $\Dij \leftarrow \Dijm + \one{\Yti \ne \Ytj},\Nij \leftarrow \Nijm + 1$
		\ENDFOR
	\end{algorithmic}
\end{algorithm}

In first round, it selects arm $K$ and initializes the value of number of comparisons and counter of disagreements for each pair $(i,j), 1 \le i<j \le K$, denoted by $\mathcal{N}_{ij}(1)$ and $\mathcal{D}_{ij}(1)$ respectively (Line $2$). In the round $t$, the algorithm computes the optimistic estimate for the disagreement probability $\opijkl(t)$ (Line $3$) using empirical estimate $\hat{p}_{ij}(t) = \frac{\Dijm}{\Nijm}$ as follows:
\begin{align}
	\label{eq:KL_Estimate}
	\opijkl(t) = \max\Big\{ &q \in \left[\hat{p}_{ij}(t), 1\right]: \Nijm d\left(\hat{p}_{ij}(t), q\right)  \nonumber \\
	&\qquad 
	\le \log(t) + a\log(\log(t)) \Big\}
\end{align}
where $d(p, q)$ denotes the Kullback-Leibler divergence between two Bernoulli distributions with parameter $p$ and $q$. 

Similar to \ref{alg:CSMD_TS}, the sets $\mathcal{B}_t^l$, $\mathcal{B}_t^h$ and their intersection (Line $5$) are computing by setting $\opij(t)$ as $\opijkl(t)$ that are used for selection of arm $I_t$ (Line $6$). After playing arm $I_t$, the feedback $\{Y^j_t\}_{j=1}^{I_t}$ (Line $7$) are observed which is used to update the values of $D_{ij}(t)$ and $N_{ij}(t)$ (Line $8$).

	\section{Best Subset Selection}
	\label{sec:subset}

Now, we consider the case where a learner can select any subsets of features which can have the smaller expected total cost compare to choosing features one after another in the fixed sequence. The learner can select any subset of features from $N_S \doteq 2^K - 1$ possible subsets (excluding empty set) of $K$ features. Each subset is indexed and corresponding prediction model is referred as an arm in multi-armed bandits problem. Let $S_t \in [N_S]$ be the index of arm selected by learner in the round $t$ and $F_t$ be the associated features with arm $S_t$. After selecting an arm $S_t$, learner incurs cost $C_{S_t} = \sum_{i \in F_t} c_i$ and misclassification penalty $\one{Y_t^{S_t} \ne \Yt}$ where $Y_t^{S_t}$ is the feedback observed for arm $S_t$. Hence,  the expected total cost incurred by the learner is $\lambda_{S_t} C_{S_t} + \gamma_{S_t}$  where $\gamma_{S_t} \doteq \Prob{Y_t^{S_t} \ne \Yt}$ is the error-rate of selecting arm $S_t$ and $\lambda_{S_t}$ normalizes error-rate and cost. WLOG, we set $\lambda_{s}= 1$ for all $s \in [N_S]$.

The goal of the learner is to find the best arm that minimizes the expected total cost. But the challenge with such problems resides in its combinatorial structure: the size of arms grows as $2^K$ where $K$ is the number of features. Such combinatorial problem is in MAB setting have been explored in \cite{ICML13_chen2013combinatorial,NIPS15_combes2015combinatorial,ICML18_wang2018thompson}. Fortunately, in medical diagnosis, the number of medical tests (features) is generally small, and hence algorithm needs to explore a small number of arms. Further, in our setup, we not only observe feedback from the selected arm but also from all arms that correspond to the subsets of associated selected arm's features. We called these arms as `basic arms' of the selected arm. Under WD property, the optimal arm $s^\star$ must lie in the following two subsets:
\begin{subequations}
	\label{eq:subsetSelection}
	\begin{align}
		&\mathcal{B}^l= \Big\{i: \forall j \in [N_S], C_{S_i} \ge C_{S_j} \ni \nonumber \\
		& \qquad \qquad \quad C_{S_i} - C_{S_j} \leq \Prob{Y^{S_{i}} \ne Y^{S_j}} \Big\} \cup \{1\} \label{set:subsetBl} \\
		&\mathcal{B}^h = \Big\{i: \forall j \in [N_S], C_{S_i} < C_{S_j} \ni \nonumber \\
		& \qquad \qquad \quad C_{S_j} - C_{S_i} > \Prob{Y^{S_{i}} \ne Y^{S_j}}\Big\} \cup \{N_S\} \label{set:subsetBh}
	\end{align}
\end{subequations}
Similar to \eqref{eq:estimateOptimal}, we replace the unknown term $\Prob{Y^{S_{i}} \ne Y^{S_j}}$ by its optimistic estimates as follows:
\begin{subequations}
	\label{eq:subsetEstOptimal}
	\begin{align}
		&\mathcal{\hat{B}}_t^l= \Big\{i: \forall j \in [N_S], C_{S_i} \ge C_{S_j} \ni \nonumber \\
		& \qquad \qquad \quad C_{S_i} - C_{S_j} \leq \opij(t) \Big\} \cup \{1\} \label{set:subsetEstBl}\\
		&\mathcal{\hat{B}}_t^h = \Big\{i: \forall j \in [N_S], C_{S_i} < C_{S_j} \ni \nonumber \\
		& \qquad \qquad \quad C_{S_j} - C_{S_i} > \opij(t) \Big\} \cup \{N_S\}. \label{set:subsetEstBh}
	\end{align}
\end{subequations}
Next, we describe UCB \cite{NIPS15_combes2015combinatorial} and Thompson Sampling \cite{ICML18_wang2018thompson} based online algorithms that check the conditions given in \eqref{eq:subsetEstOptimal} for selecting the optimal arm.

\vspace{-0.05cm}
\subsection{ESCB based Algorithm: \ref{alg:CSMD_ESCB}}
A UCB based algorithm ESCB (Efficient Sampling for Combinatorial Bandits) for solving combinatorial bandits is proposed in \cite{NIPS15_combes2015combinatorial}, which treat each subset as an arm. We adapt ESCB for solving the combinatorial CSMD problem and develop an algorithm named \ref{alg:CSMD_ESCB}. This algorithm works similar to \ref{alg:CSMD_KL}, except statistics updates which are changed only for  `basic arms' of the selected arm (Line $10$). Similar to ESCB, \ref{alg:CSMD_ESCB} uses two UCB based indices for computing the optimistic estimate of the disagreement probability (Line $4$). One is the kl-UCB as given in the equation \eqref{eq:KL_Estimate} and another index is based on UCB1 as given by the following equation for the round $t$: 
\begin{equation}
	\label{eq:updateByUCB}
	\opijucb(t) = \frac{\Dijm}{\Nijm} + \sqrt{\frac{\alpha\log t}{\Nijm}}.
\end{equation}

\begin{algorithm}[!t]
	\renewcommand{\thealgorithm}{\textnormal{CSMD-ESCB}}
	\floatname{algorithm}{}
	\caption{Algorithm for \textbf{CSMD} problem using \textbf{ESCB} }
	\label{alg:CSMD_ESCB}
	\begin{algorithmic}[1]
		\STATE \textbf{Input:}  $a \ge 0$ (kl-UCB index) or $\alpha>0.5$ (UCB1 index)
		\STATE Select arm $N_S = 2^K - 1$ and observe $Y^1,\dots,Y^{N_S}$
		\STATE Set $\forall 1 \le i< j \le N_S: \mathcal{D}_{ij}(1) \leftarrow \one{Y^i_1 \ne Y^j_1}, \mathcal{N}_{ij}(1) \leftarrow 1$
		\FOR{$t=2,3,...$}
			\STATE $\forall 1 \le i< j \le N_S:$ compute $\opijkl(t)$ using \eqref{eq:KL_Estimate} (kl-UCB index) or $\opijucb1(t)$ using \eqref{eq:updateByUCB} (UCB1 index)
			\STATE Compute $\mathcal{\hat{B}}_t^l$ and $\mathcal{\hat{B}}_t^h$ as given in \eqref{set:subsetEstBl} and \eqref{set:subsetEstBh} by setting $\opij(t)= \opijkl(t)$ or $\opijucb1(t)$. Set $\mathcal{\hat{B}}_t := \mathcal{\hat{B}}_t^l \cap \mathcal{\hat{B}}_t^h$
			\STATE $S_t\leftarrow \min \left\{\mathcal{\hat{B}}_t\cup \{N_S\} \right\}$
			\STATE $P_I(S_t) \leftarrow$ indices of basic arms of selected arm $S_t$
			\STATE Select arm $S_t$ and observe $\{Y_t^s\}_{s \in P_I(S_t)}$
			\STATE $\forall i,j \in P_I(S_t), i < j:$ update $\Dij \leftarrow \Dijm + \one{\Yti \ne \Ytj},\Nij \leftarrow \Nijm + 1$
		\ENDFOR
	\end{algorithmic}
\end{algorithm}

\subsection{Combinatorial TS\ based Algorithm: \ref{alg:CSMD_CTS}}
Thompson Sampling is used for combinatorial bandits problem \cite{ICML18_wang2018thompson}. We also develop a Thompson Sampling based algorithm for combinatorial CMED problem, named \ref{alg:CSMD_CTS}. This algorithm works like \ref{alg:CSMD_TS}, except the statistics updates which are changed only for  `basic arms' of the selected arm (Line $7$).
\begin{algorithm}[!t]
	\renewcommand{\thealgorithm}{\textnormal{CSMD-CTS}}
	\floatname{algorithm}{}
	\caption{Algorithm for \textbf{CSMD} problem using \textbf{CTS}} 
	\label{alg:CSMD_CTS}
	\begin{algorithmic}[1]
		\STATE Set $ \forall 1 \le i< j \le N_S: \mathcal{S}_{ij}(1) \leftarrow 1, \mathcal{F}_{ij}(1) \leftarrow 1$
		\FOR{$t=1,2,...$}
			\STATE $\forall 1 \le i< j \le N_S: \opijts(t) \leftarrow \mbox{Beta}(\mathcal{S}_{ij}(t), \mathcal{F}_{ij}(t))$
			\STATE Compute $\mathcal{\hat{B}}_t^l$ and $\mathcal{\hat{B}}_t^h$ as given in \eqref{set:subsetEstBl} and \eqref{set:subsetEstBh} by setting $\opij(t) = \opijts(t)$. Set $\mathcal{\hat{B}}_t := \mathcal{\hat{B}}_t^l \cap \mathcal{\hat{B}}_t^h$
			\STATE $S_t\leftarrow \min \left\{\mathcal{\hat{B}}_t\cup \{N_S\}\right\}$
			\STATE $P_I(S_t) \leftarrow$  indices of basic arms of selected arm $S_t$
			\STATE Select arm $S_t$ and observe $\{Y_t^s\}_{s \in P_I(S_t)}$
			\STATE $\forall i,j \in P_I(S_t), i <j:$ update $\mathcal{S}_{ij}(t+1) \leftarrow\mathcal{S}_{ij}(t) + \one{\Yti \ne \Ytj}, \mathcal{F}_{ij}(t+1) \leftarrow\mathcal{F}_{ij}(t) + \one{\Yti = \Ytj}$
		\ENDFOR
	\end{algorithmic}
\end{algorithm}

	\section{Experiments}
	\label{sec:experiments}

We empirically evaluate the performance of the proposed algorithms on different problem instances derived from two real healthcare datasets: Heart Disease (Cleveland) \cite{HEART98_robert1988va} and PIMA Indians Diabetes \cite{UCI16_pima2016kaggale}. We used these datasets as the cost of acquiring individual features is specified. In our experiments, the prediction model corresponding to each arm is a binary classifier. We repeat each experiment 100 times and present the cumulative regret with a 95\% confidence interval. The vertical line on each curve shows the confidence interval. 

\textbf{Details of the used problem instances:} We split the features of both datasets into three subsets based on their costs where each subset is equivalent to a test. Then we train linear classifiers on these subsets using logistic regression. For Heart Disease dataset, we associate 1st classifier with the first seven attributes that include cholesterol readings, blood-sugar, and rest-ECG, which costs \$$32$). The 2nd classifier also utilizes the thalach, exang, and oldpeak attributes that cost \$$397$, and the 3rd classifier utilizes more extensive tests at a total cost of \$$601$. For the PIMA-Diabetes dataset, the first classifier is associated with patient history/profile at the cost of \$$6$, the 2nd classifier, also, utilizes glucose tolerance test (cost \$$29$) and the 3rd classifier uses all attributes including insulin test (cost \$$46$). Since the costs of features are all greater than one, we normalize costs using a tuning parameter $\lambda$ (as defined in Section 2). In our setup, high (low)-values for $\lambda$ correspond to low (high)-budget constraint. For example, if we set a fixed budget of \$100, this corresponds to high-budget (small $\lambda$) and low budget (large $\lambda$) for PIMA Indians Diabetes (3rd classifier optimal) and Heart Disease (1st classifier optimal) respectively. For performance evaluation, different values of $\lambda$ are used in five problem instances for both real datasets, as given in Table \ref{table:real_datasets} which are taken from \cite{AISTATS19_verma2019online}. As the size of both datasets is small: Heart Disease dataset (\# of samples=$297$) and PIMA Indians Diabetes (\# of samples=$768$), we use random over-sampling to increase the sample size to $10000$.

\begin{table*}
	\centering
	\small
	\caption{Different parameters of problem instances which are derived from real datasets.	Note that WD doesn't hold for Case 5 and $\lambda$ value of optimal classifier is in \textbf{bold} font.}
	\label{table:real_datasets}
	\setlength\tabcolsep{10pt}
	\setlength\extrarowheight{1pt}
	\begin{tabularx}{0.725\textwidth}{|p{1.43cm}|p{0.82cm}|p{0.85cm}|p{0.91cm}|p{0.96cm}|p{0.96cm}|p{0.98cm}|p{0.48cm}|}
		\hline
		\multirow{2}{*}{\parbox{2cm}{Values/ \newline Classifiers}} &\multicolumn{3}{|c|}{\bf PIMA-Diabetes}&\multicolumn{3}{|c|}{\bf Heart Disease} &\multicolumn{1}{|c|}{\multirow{4}{*}{\parbox{.48cm}{WD Pro.}}}\\ \cline{2-7} 
		&Clf. 1 &Clf. 2&Clf. 3&Clf. 1 &Clf. 2&Clf. 3&\\ \cline{1-7}
		Error-rate & 0.3125 &0.2331&0.2279&0.29292 &0.20202& 0.14815 &\\ \cline{1-7}
		Cost (in \$) & 4 &29&46&32 &397 &601&\\ 
		\hline
		$\lambda$ in Case  1& \textcolor{black}{\textbf{0.01}}& 0.0106& 0.015&\textcolor{black}{\textbf{0.0001}}& 0.0008& 0.001&\multicolumn{1}{|c|}{\checkmark}\\ 
		\hline
		$\lambda$ in Case  2& 0.01& \textcolor{black}{\textbf{0.004}}& 0.0038&0.0001& \textcolor{black}{\textbf{0.0001}}& 0.00035&\multicolumn{1}{|c|}{\checkmark}\\ 
		\hline
		$\lambda$ in Case  3& \textcolor{black}{\textbf{0.01}}& 0.0113& 0.015&\textcolor{black}{\textbf{0.0001}}& 0.0009& 0.001&\multicolumn{1}{|c|}{\checkmark}\\ 
		\hline
		$\lambda$ in Case  4& 0.0001& 0.0001& \textcolor{black}{\textbf{0.0001}}&0.00001& 0.00004& \textcolor{black}{\textbf{0.0001}}&\multicolumn{1}{|c|}{\checkmark}\\ 
		\hline
		$\lambda$ in Case  5& 0.01& \textcolor{black}{\textbf{0.002}}& 0.0055&0.0042& \textcolor{black}{\textbf{0.0001}}& 0.00027&\multicolumn{1}{|c|}{\ding{53}}\\ 
		\hline
	\end{tabularx}
\end{table*}


\begin{figure*}[!t]
	\centering
	\begin{subfigure}[b]{0.2455\textwidth}
		\includegraphics[width=\linewidth]{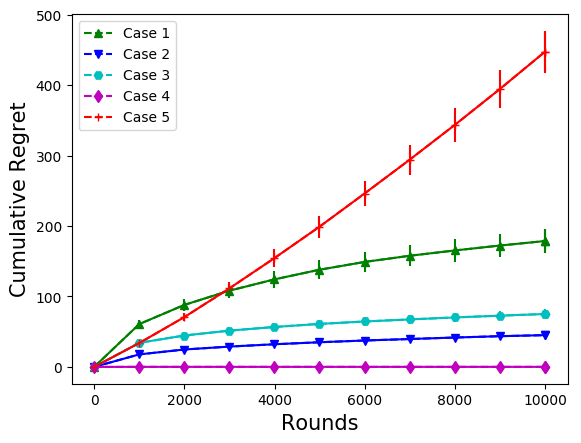}
		\caption{\footnotesize \ref{alg:CSMD_TS}}
		\label{fig:heartCMED-TS}
	\end{subfigure}
	\begin{subfigure}[b]{0.2455\textwidth}
		\includegraphics[width=\linewidth]{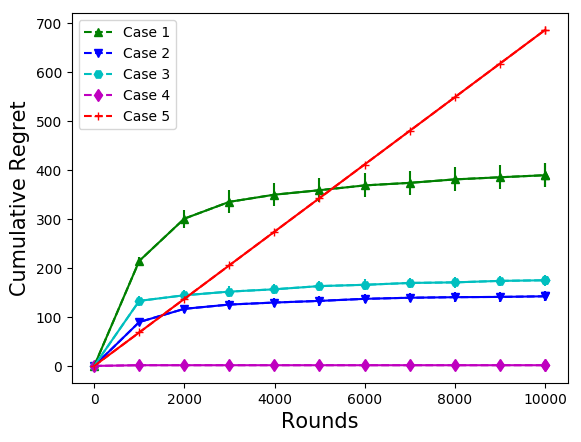}
		\caption{\footnotesize \ref{alg:CSMD_KL} (with $c=0$)}
		\label{fig:heartCMED-KL}	
	\end{subfigure}
	\begin{subfigure}[b]{0.2455\textwidth}
		\includegraphics[width=\linewidth]{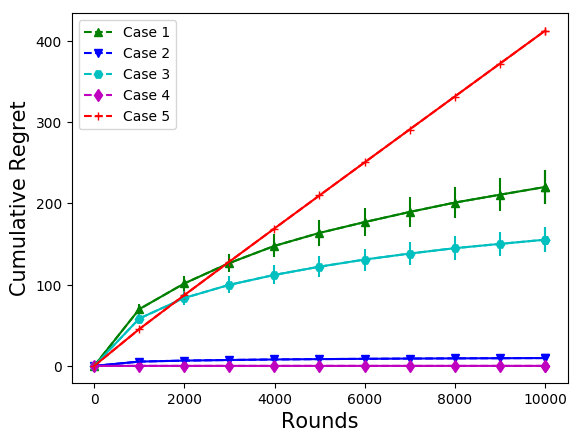}
		\caption{\footnotesize \ref{alg:CSMD_TS}}
		\label{fig:diabetesCMED-TS}
	\end{subfigure}
	\begin{subfigure}[b]{0.2455\textwidth}
		\includegraphics[width=\linewidth]{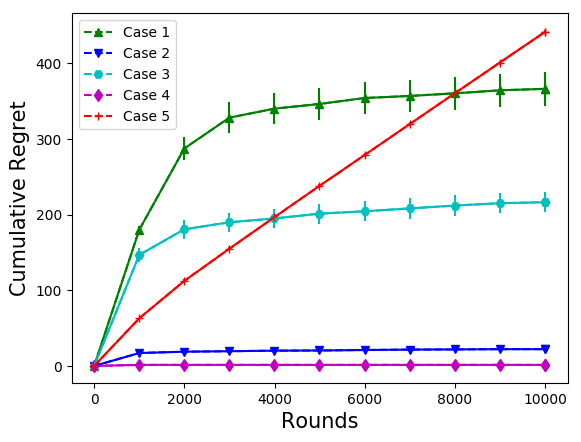}
		\caption{\footnotesize \ref{alg:CSMD_KL} (with $c=0$)}
		\label{fig:diabetesCMED-KL}	
	\end{subfigure}
	\caption{\small Cumulative regret of \ref{alg:CSMD_TS} and \ref{alg:CSMD_KL} with $c=0$ for different problem instances of Heart Disease dataset (Fig. \eqref{fig:heartCMED-TS} and \eqref{fig:heartCMED-KL}) and PIMA Indians Diabetes dataset (Fig. \eqref{fig:diabetesCMED-TS} and \eqref{fig:diabetesCMED-KL}).}
	\label{fig:cascade}
\end{figure*}
\begin{figure*}[!t]
	\centering
	\begin{subfigure}[b]{0.2455\textwidth}
		\includegraphics[width=\linewidth]{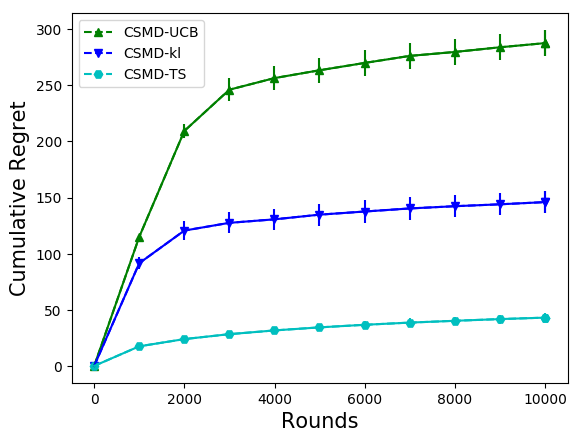}
		\caption{\footnotesize Heart Disease (Case 1)}
		\label{fig:compareHeart}	
	\end{subfigure}
	\begin{subfigure}[b]{0.2455\textwidth}
		\includegraphics[width=\linewidth]{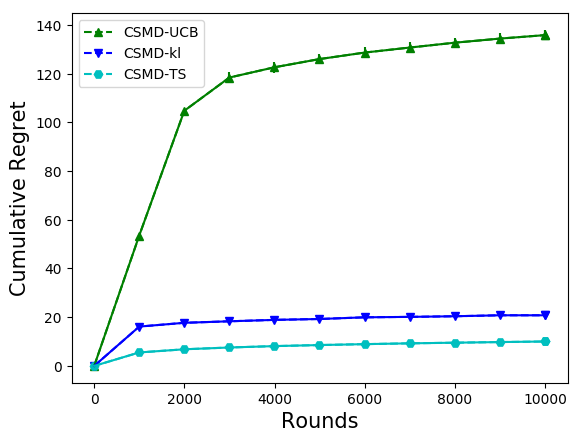}
		\caption{\footnotesize PIMA Indians Diabetes (Case 1)}
		\label{fig:compareDiabetes}
	\end{subfigure}
	\begin{subfigure}[b]{0.2455\textwidth}
		\includegraphics[width=\linewidth]{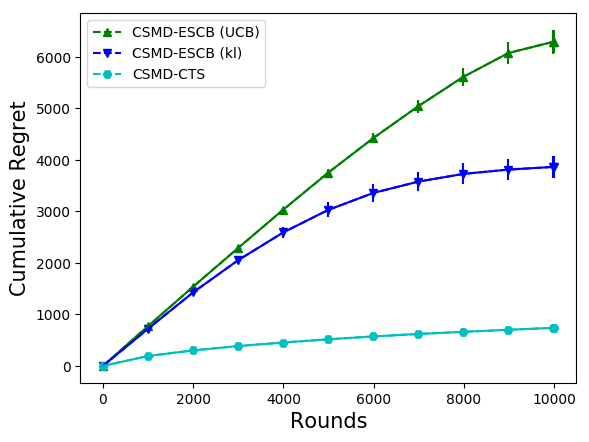}
		\caption{\footnotesize Heart Disease}
		\label{fig:compareSubHeart}	
	\end{subfigure}
	\begin{subfigure}[b]{0.2455\textwidth}
		\includegraphics[width=\linewidth]{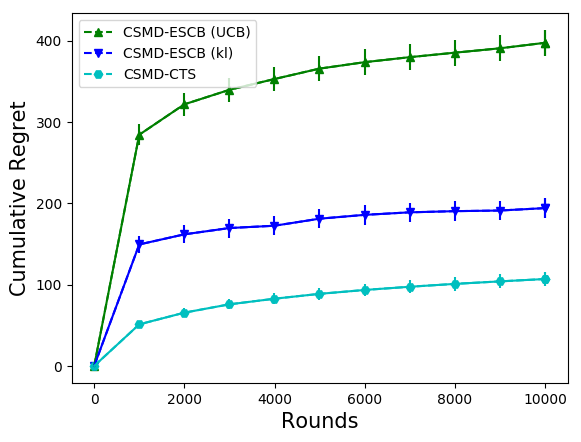}
		\caption{\footnotesize PIMA Indians Diabetes}
		\label{fig:compareSubDiabetes}
	\end{subfigure}
	\caption{\small Comparison of cumulative regret of different algorithms for CSMD problem with assumtion that cheaper tests are selected before expensive tests (Fig. \eqref{fig:compareHeart} and \eqref{fig:compareDiabetes}) and combinatorial CSMD problem (Fig. \eqref{fig:compareSubHeart} and \ref{fig:compareSubDiabetes}).}
	\label{fig:subset}
\end{figure*}

\textbf{Cumulative regret v/s different $\lambda$ values:} We evaluate our algorithms on different problem instances where each problem instance has different values of $\lambda$. Only the problem instances that satisfy WD property have sub-linear regret as shown in Figure \ref{fig:cascade} for Heart Disease and PIMA Indians Diabetes datasets. The regret depends on how well the WD property is satisfied by problem instance. The wellness is measured in term of $\xi$ (as defined in \eqref{eq:WD}) and the larger the value of $\xi$, smaller will be the regret and vice-versa. Note that the problem instance $5$ does not satisfy WD property and has a linear regret.

\textbf{Comparison between different algorithms:} We compare the performance of \ref{alg:CSMD_TS}, \ref{alg:CSMD_KL} and UCB1 index based algorithm CSMD-UCB on Heart Disease and PIMA Indians Diabetes datasets. CSMD-UCB is motivated from USS-UCB algorithm of \cite{AISTATS19_verma2019online}. It works similar to \ref{alg:CSMD_KL} except that the optimistic estimates are computed using \eqref{eq:updateByUCB} with value of $\alpha=0.51$. As expected, under cascade condition where a feature is selected in a fixed sequence, \ref{alg:CSMD_TS} outperforms \ref{alg:CSMD_KL} and CSMD-UCB as shown in Figure \ref{fig:subset}. We relax cascade condition in Section 4 and propose two new algorithms: \ref{alg:CSMD_CTS} and \ref{alg:CSMD_ESCB} for solving combinatorial CSMD problem.  As depicted in Figure \ref{fig:subset}, again as expected, Thompson Sampling based algorithm \ref{alg:CSMD_CTS} outperforms both kl-UCB based algorithm and UCB1 index based algorithms.

	\section{Conclusion}
	\label{sec:conclusion}

We model Cost-Sensitive Medical Diagnosis (CSMD) as an online feature selection problem where both accuracy and cost of tests are important. Due to the unavailability of true states, the setting is unsupervised, but the tests' predictions can be observed. The tests form a cascade and are ordered by their increasing cost. First, we consider a setting where a test can be used only after all previous tests in the cascade are done and propose two algorithms for solving it. Later we relax the condition and allow the learner to pick any subset that minimizes the overall total cost. The size of such a subset makes it a combinatorial problem. We propose Thompson Sampling and UCB index based algorithms for solving such combinatorial CSMD problem. We demonstrate the performance of our algorithms on two real datasets and empirically show that any problem instance satisfying WD property has sub-linear regret. As future work, we would like to derive the regret bounds for the proposed algorithms.

	\section*{Acknowledgments}
	Manjesh K. Hanawal would like to thank the support from INSPIRE faculty fellowships from DST, Government of India, SEED grant (16IRCCSG010) from IIT Bombay, and Early Career Research (ECR) Award from SERB.

	\bibliographystyle{IEEEtran}
	\bibliography{ref}

\end{document}